\documentclass[times,twocolumn,final,authoryear]{elsarticle}

\usepackage{ycviu}
\usepackage{framed,multirow}

\usepackage{amssymb}
\usepackage{latexsym}

\usepackage{url}
\usepackage{xcolor}
\definecolor{newcolor}{rgb}{.8,.349,.1}

\journal{Computer Vision and Image Understanding}

\begin{document}

\ifpreprint
  \setcounter{page}{1}
\else
  \setcounter{page}{1}
\fi

\begin{frontmatter}

\title{SalGAN: visual saliency prediction with adversarial networks}

\author[1]{Junting \snm{Pan}}
\author[2]{Cristian \snm{Canton-Ferrer}}
\author[3]{Kevin \snm{McGuinness}}
\author[3]{Noel E. \snm{O'Connor}}
\author[4]{Jordi \snm{Torres}}
\author[1]{Elisa \snm{Sayrol}}
\author[1]{Xavier \snm{Giro-i-Nieto}\corref{cor1}} 
\cortext[cor1]{Corresponding author: 
  Tel.: +34-934-015-769;}
\ead{xavier.giro@upc.edu}

\address[1]{Universitat Politecnica de Catalunya, Barcelona 08034, Catalonia/Spain}
\address[2]{Facebook AML, Seattle, WA, United States of America}
\address[3]{Insight Center for Data Analytics, Dublin City University, Dublin 9, Ireland}
\address[4]{Barcelona Supercomputing Center, Barcelona 08034, Catalonia/Spain}


\begin{abstract}
Recent approaches for saliency prediction are generally trained with a loss function based on a single saliency metric.  This could lead to low  performance when evaluating with other saliency metrics.  
In this paper, we propose a novel data-driven metric based saliency prediction method, named SalGAN (Saliency GAN), trained with adversarial loss function. SalGAN consists of two networks: one predicts saliency maps from raw pixels of an input image; the other one takes the output of the first one to discriminate whether a saliency map is a predicted one or ground truth.
By trying to make the predicted saliency map indistinguishable with the ground truth, SalGAN is expected to generate saliency maps that resembles the ground truth. Our experiments show that the adversarial training allows our model to obtain state-of-the-art performances across various saliency metrics.
\end{abstract}


\end{frontmatter}


\section{Introduction}
\label{sec:Motivation}

Visual saliency describes the spatial locations in an image that attract the human attention.  It is understood as a result of a bottom-up process where a human observer explores the image for a few seconds with no particular task in mind. Therefore, saliency prediction is indispensable for various machine vision tasks such as object recognition~\citep{walther2002attentional}.

Visual saliency data are traditionally collected by eye-trackers~\citep{judd2009learning}, and more recently with mouse clicks \citep{jiang2015salicon} or webcams \citep{cvpr2016_Khosla}. The salient points of the image are aggregated and convolved with a Gaussian kernel to obtain a saliency map. As a result, a gray-scale image or a heat map is generated to represent the probability of each corresponding pixel in the image to capture the human attention.


\begin{figure}
\centering
\begin{tabular}{cc}
Image & Ground Truth \\
\includegraphics[width=0.4\columnwidth]{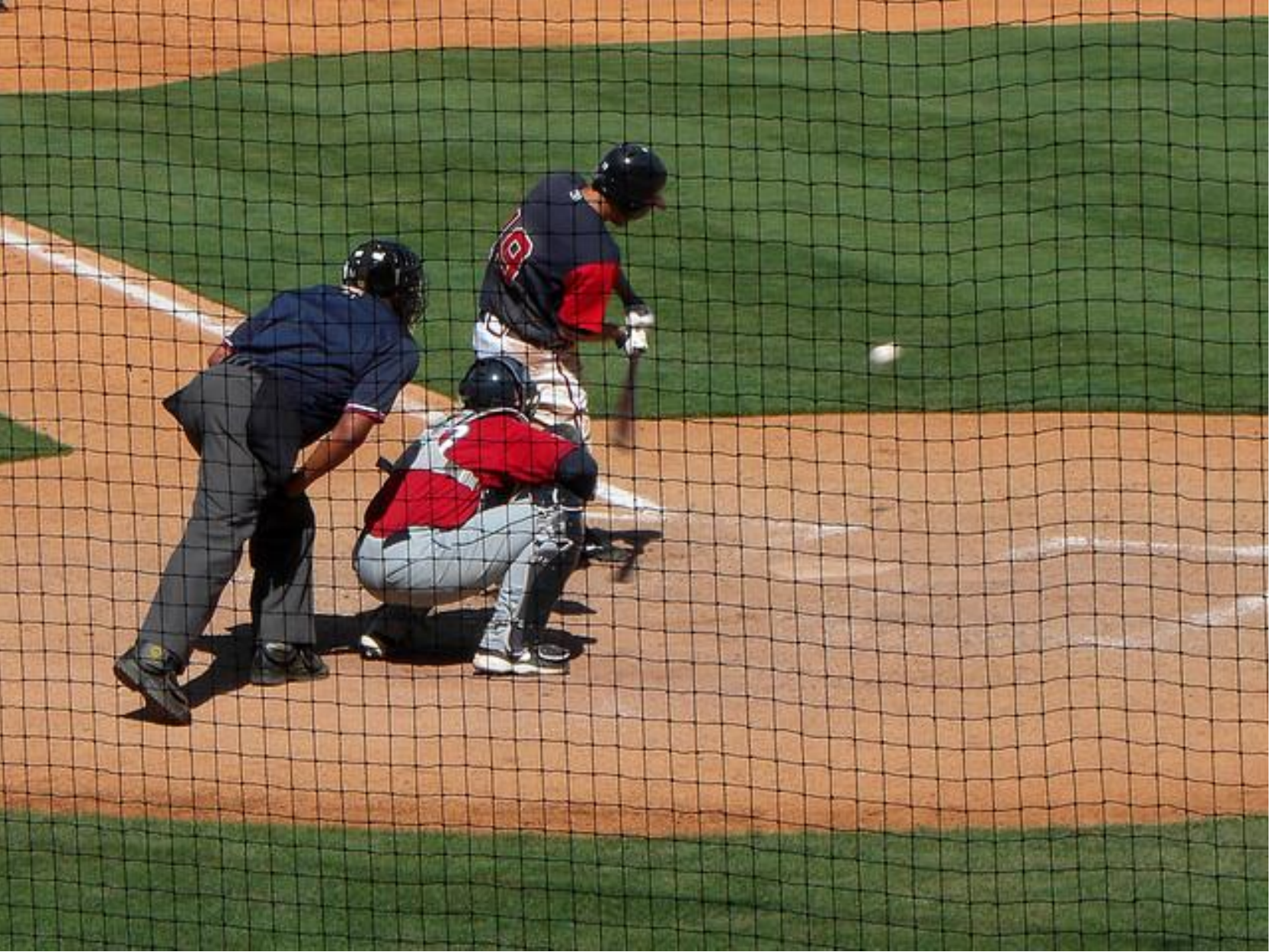} &     \includegraphics[width=0.4\columnwidth]{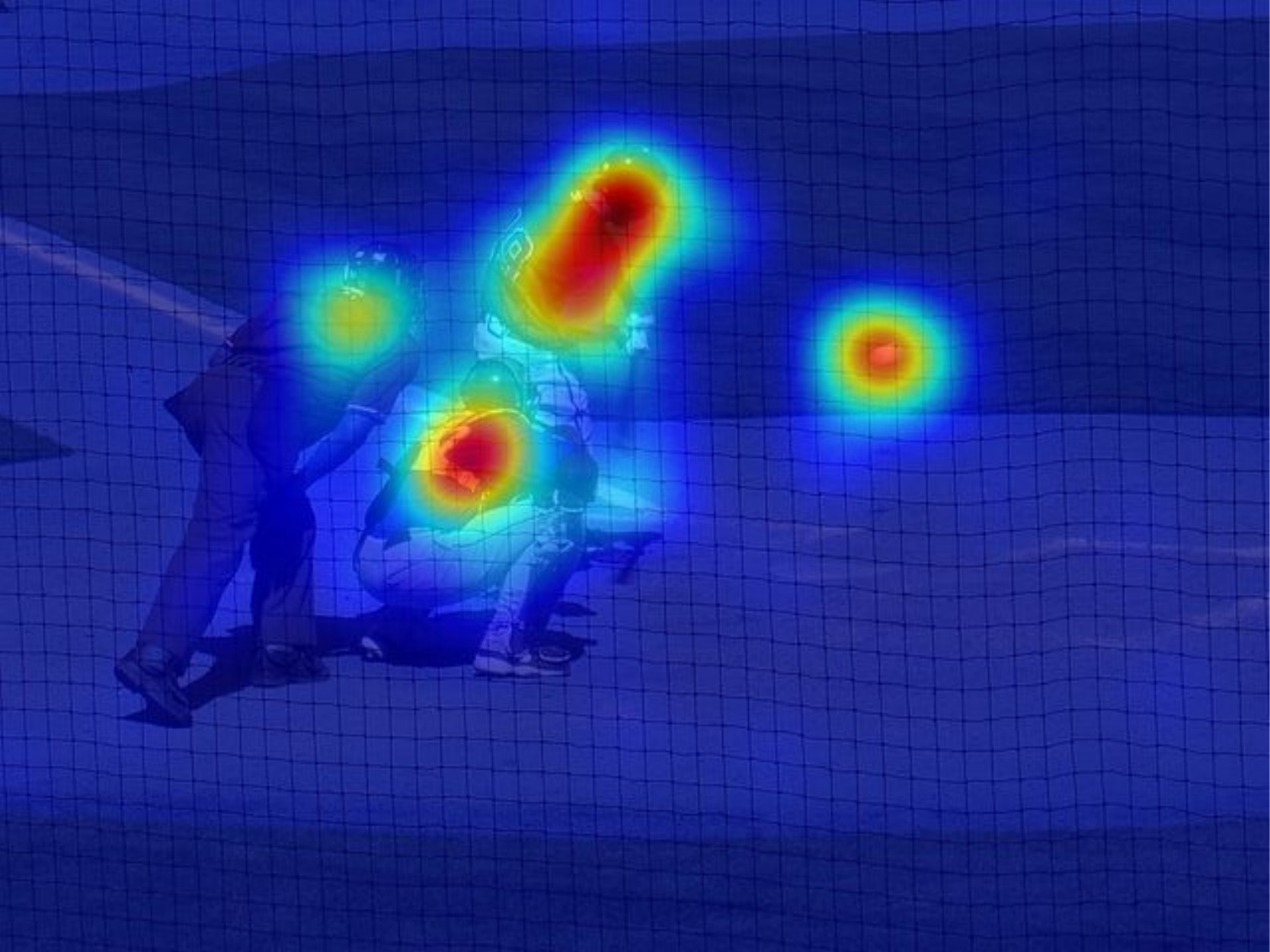} \\
BCE & SalGAN  \\
\includegraphics[width=0.4\columnwidth]{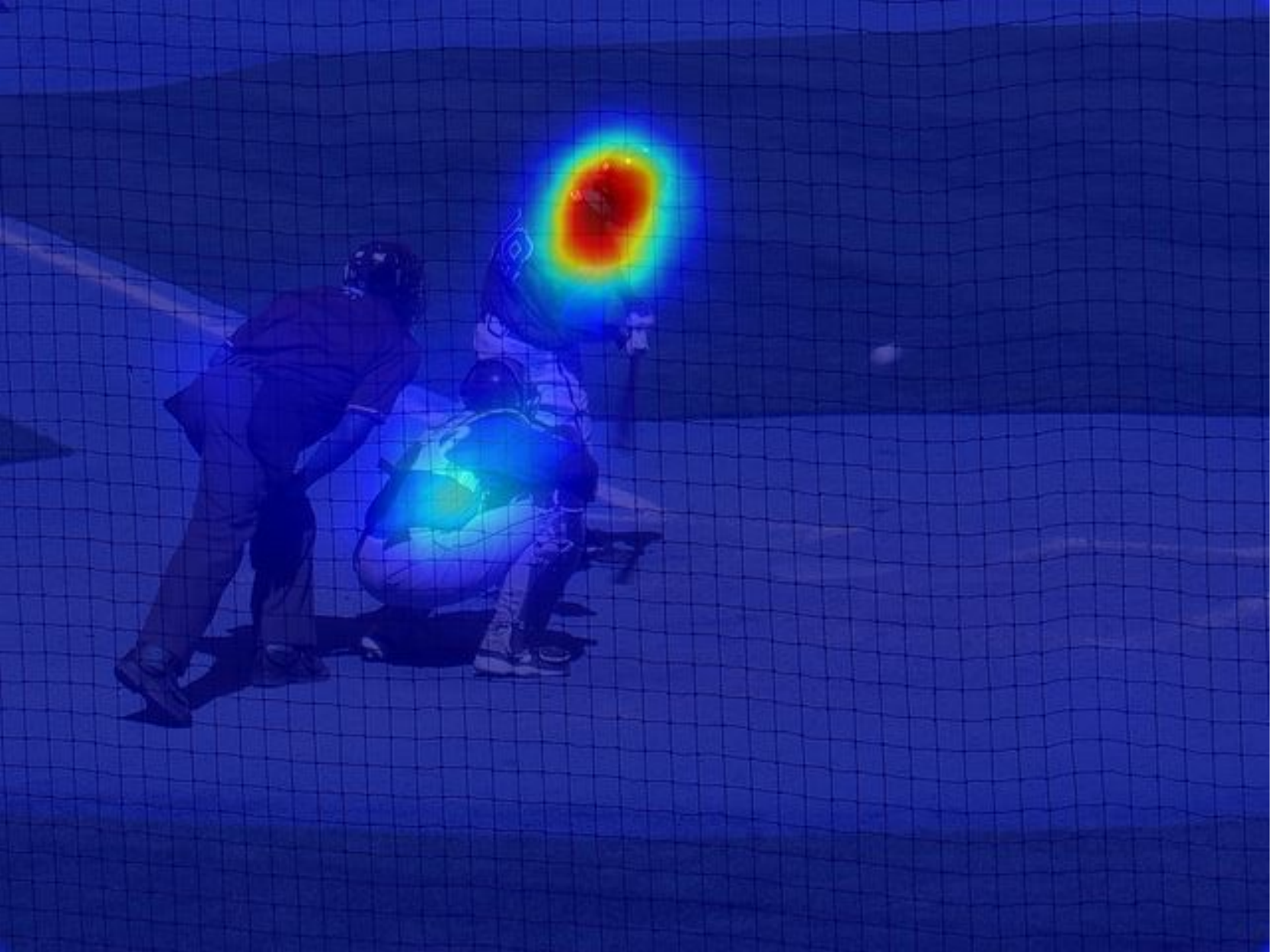} &     \includegraphics[width=0.4\columnwidth]{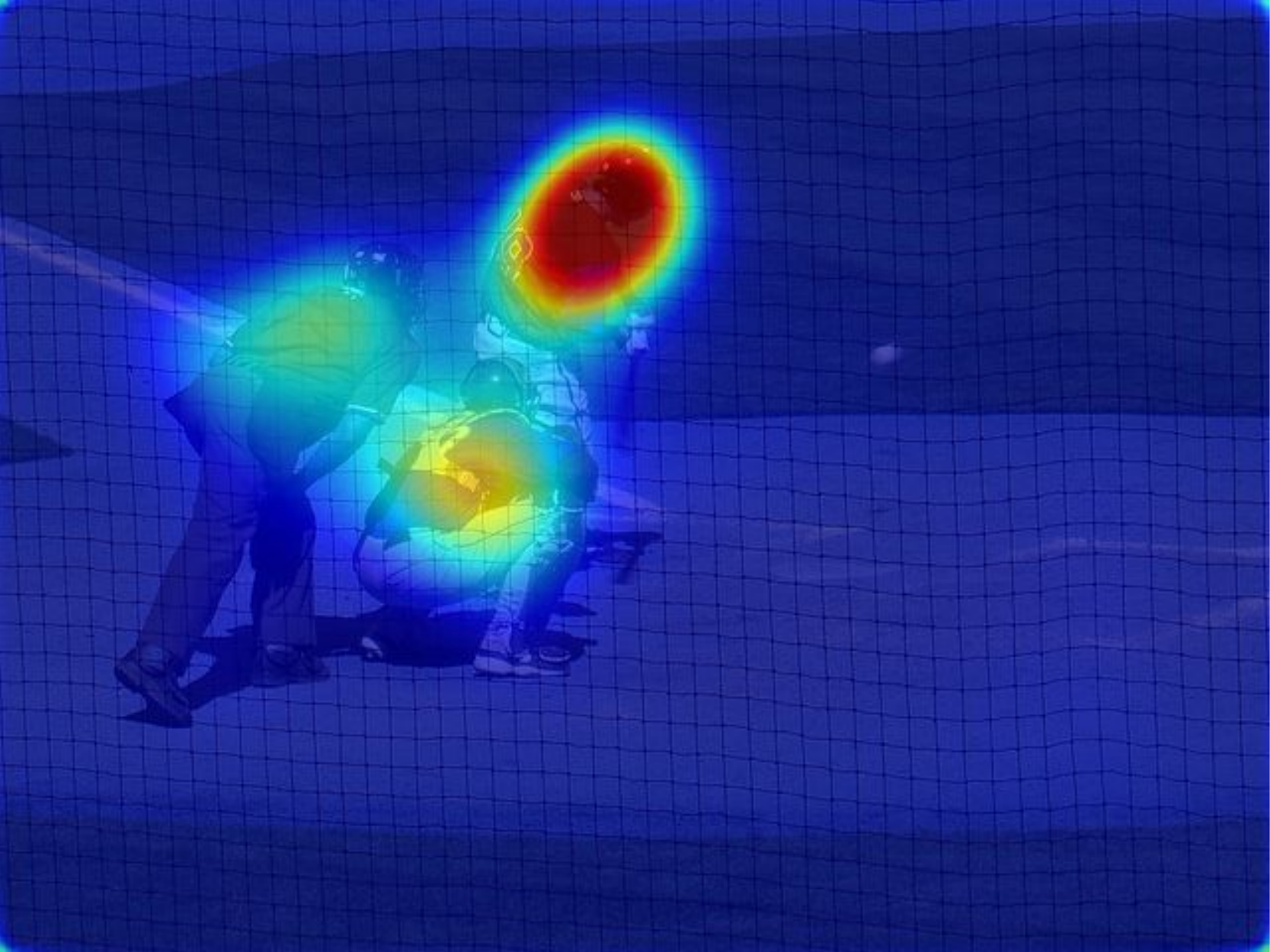} \\
\end{tabular}
\caption{Example of saliency map generation where the proposed system (SalGAN) outperforms a standard binary cross entropy (BCE) prediction model.}
\label{fig:catchyFigure}
\end{figure}
A lot of research effort has been made in designing an optimal loss function for saliency prediction. State-of-the-art methods \citep{huang2015salicon} adopt saliency based metrics while others \citep{Pan_2016_CVPR,mlnet2016,jetley2016end,sun2017integrated} use distance in saliency map space. How to choose or design a best training loss is still an open problem. In addition, different saliency metrics diverge at defining the meaning of saliency maps, and there exist inconsistency with the model comparison. For instance, it has been pointed that the optimal metric for model optimization may depend on the final application  \citep{Bylinskii2016metrics}.


To this end, instead of designing  a tailored loss function, we introduce adversarial training for visual saliency prediction inspired by generative adversarial networks (GANs)\citep{goodfellow2014generative}. We dub the proposed method as SalGAN. We focus on exploring the benefits of using such an adversarial loss to make the output saliency map not able to be distinguished from the real saliency maps. 
In GANs, training is driven by two competing agents: first, the \textit{generator} synthesizing samples that match with the training data; second, the \textit{discriminator} distinguishing between a real sample drawn directly from the training data and a fake one synthesized by the generator. In our case, this data distribution corresponds to pairs of real images and their corresponding visual saliency maps. 

\begin{figure*}[!ht]
\includegraphics[width=\textwidth]{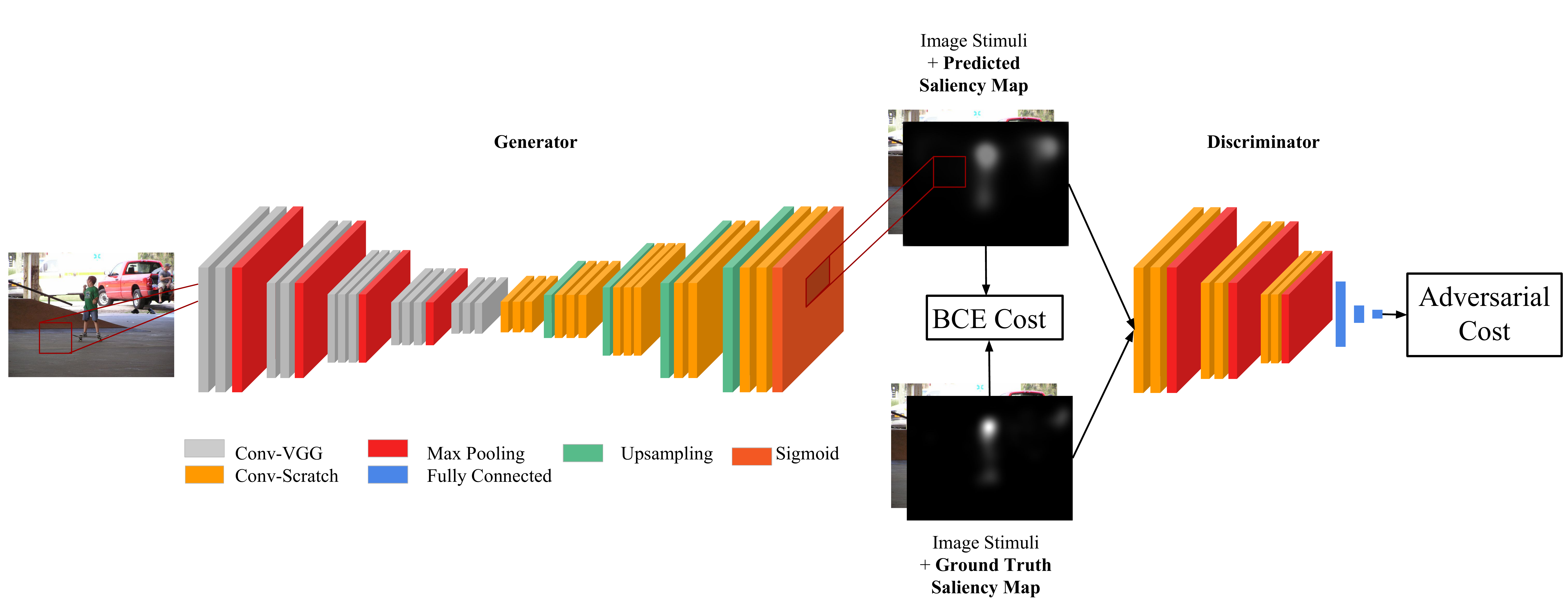}
 \caption{Overall architecture of the proposed saliency system. The input for the saliency prediction network predicts an output saliency map given a natural image as input. Then, the pair of saliency and image is feed into the discriminator network. The output of the discriminator is a score that tells about whether the input saliency map is real or fake.} 
\label{fig:architecture}
\end{figure*}

Specifically, SalGAN estimates the saliency map of an input image using a deep convolutional neural network (DCNN). As shown in the Figure~\ref{fig:architecture} this network is initially trained with a binary cross entropy (BCE) loss over down-sampled versions of the saliency maps. The model is then refined with a discriminator network trained to solve a binary classification task between the saliency maps generated by SalGAN and the real ones used as ground truth. Our experiments show how adversarial training allows reaching state-of-the-art performance across different metrics when combined with a BCE content loss in a single-tower and single-task model. 


To summarize, we investigate the introduction of the adversarial loss to the visual saliency learning. By introducing adversarial loss to the BCE saliency prediction model, we achieve the state-of-the-art performance in MIT300 and SALICON dataset for almost all the evaluation metrics

The remaining of the text is organized as follows. 
Section \ref{sec:RelatedWork} reviews the state-of-the-art models for visual saliency prediction, discussing the loss functions they are based upon, their relations with the different metrics as well as their complexity in terms of architecture and training. 
Section \ref{sec:Architecture} presents SalGAN, our deep convolutional neural network based on a convolutional encoder-decoder architecture, as well as the discriminator network used during its adversarial training.
Section \ref{sec:Training} describes the training process of SalGAN and the loss functions used.
Section \ref{sec:Experiments} includes the experiments and results of the presented techniques. Finally, Section \ref{sec:Conclusions} closes the paper by drawing the main conclusions.

Our results can be reproduced with the source code
and trained models available at \url{https://imatge-upc.github.io/saliency-salgan-2017/}.

\section{Related work}
\label{sec:RelatedWork}

Saliency prediction has received interest by the research community for many years. Thus seminal works \citep{Itti1998PAMI} proposed to predict saliency maps considering low-level features at multiple scales and combining them to form a saliency map. \citep{harel2006nips}, also starting from low-level feature maps, introduced a graph-based saliency model that defines Markov chains over various image maps, and treat the equilibrium distribution over map locations as activation and saliency values. \citep{judd2009iccv} presented a bottom-up, top-down model of saliency based not only on low but mid and high-level image features. \citep{borji2012cvpr} combined low-level features saliency maps of previous best bottom-up models with top-down cognitive visual features and learned a direct mapping from those features to eye fixations.

As in many other fields in computer vision, a number of deep learning solutions have very recently been proposed that significantly improve the performance.
For example, the Ensemble of Deep Networks (eDN) \citep{vig2014large} represented an early architecture that automatically learns the representations for saliency prediction, blending feature maps from different layers. 
In \citep{Pan_2016_CVPR} two convolutional neural networks trained ebd-to-end for saliency prediction are compared, a lighter one designed and trained from scratch, and a second and deeper one pre-trained for image classification. DCNN have shown better results even when pre-trained with datasets build for other purposes. DeepGaze \cite{kummerer2015deep} provided a deeper network using the well-know AlexNet \citep{krizhevsky2012imagenet}, with pre-trained weights on Imagenet \citep{deng2009imagenet} and with a readout network on top whose inputs consisted of some layer outputs of AlexNet. The output of the network is blurred, center biased and converted to a probability distribution using a softmax. Huang et al.~\citep{huang2015salicon}, in the so call SALICON net, obtained better results by using VGG rather than AlexNet or GoogleNet \citep{Szegedy2015}. In their proposal they considered two networks with fine and coarse inputs, whose feature maps outputs are concatenated.

Li et al. \citep{Li_2015_CVPR} proposed a multi resolution convolutional neural network that is trained from image regions centered on fixation and non-fixation locations over multiple  resolutions. Diverse top-down visual features can be learned in higher layers and  bottom-up visual saliency can also be inferred by combining information over multiple resolutions. 
These ideas are further developed in they recent work called DSCLRCN \cite{liu2018deep}, where the proposed model learns saliency related local features on each image location in parallel and then learns to simultaneously incorporate global context and scene context to infer saliency. They incorporate a model to effectively learn long-term spatial interactions and scene contextual modulation to infer image saliency. 
Deep Gaze II \citep{kummerer2017understanding} sets the state of the art in the MIT300 dataset by combining features trained for image recognition with four layer of 1x1 convolutions. 
Both DSCLRCN and Deep Gaze II obtain excellent results in the benchmarks when combined with a center bias, which is not considered in SalGAN as the results are purely the results at inference time.
MLNET \citep{mlnet2016} proposes an architecture that combines features extracted at different levels of a DCNN. They introduce a loss function inspired by three objectives: to measure similarity with the ground truth, to keep invariance of predictive maps to their maximum and to give importance to pixels with high ground truth fixation probability. In fact choosing an appropriate loss function has become an issue that can lead to improved results. Thus, another interesting contribution of ~\citep{huang2015salicon} lies on minimizing loss functions based on metrics that are differentiable, such as NSS, CC, SIM and KL divergence to train the network (see \cite{riche2013iccv} and  \cite{kummerer2054IG} for the definition of these metrics. A thorough comparison of metrics can be found in \citep{Bylinskii2016metrics}). In \citep{huang2015salicon} KL divergence gave the best results. \citep{jetley2016end} also tested loss functions based on probability distances, such as {X2} divergence, total variation distance, KL divergence and Bhattacharyya distance by considering saliency map models as generalized Bernoulli distributions. The Bhattacharyya distance was found to give the best results. 

In our work we present a network architecture that takes a different approach. By incorporating the high-level adversarial loss into the conventional saliency prediction training approach, the proposed method achieves the state-of-the-art performance in both MIT300 and SALICON datasets by a clear margin.

\section{Architecture}
\label{sec:Architecture}

The training of SalGAN is the result of two competing convolutional neural networks: a generator of saliency maps, which is SalGAN itself, and a discriminator network, which aims at distinguishing between the real saliency maps and those generated by SalGAN. This section provides details on the structure of both modules, the considered loss functions, and the initialization before beginning adversarial training.
Figure~~\ref{fig:architecture} shows the architecture of the system.

\subsection{Generator}
\label{ssec:Saliency Prediction Network}

The generator network, SalGAN, adopts a convolutional encoder-decoder architecture, where the encoder part includes max pooling layers that decrease the size of the feature maps, while the decoder part uses upsampling layers followed by convolutional filters to construct an output that is the same resolution as the input. 

The encoder part of the network is identical in architecture to VGG-16 \citep{simonyan2014very}, omitting the final pooling and fully connected layers. The network is initialized with the weights of a VGG-16 model trained on the ImageNet data set for object classification \citep{deng2009imagenet}. Only the last two groups of convolutional layers in VGG-16 are modified during the training for saliency prediction, while the earlier layers remain fixed from the original VGG-16 model. We fix weights to save computational resources during training, even at the possible expense of some loss in performance.

The decoder architecture is structured in the same way as the encoder, but with the ordering of layers reversed, and with pooling layers being replaced by upsampling layers. Again, ReLU non-linearities are used in all convolution layers, and a final $1\times 1$ convolution layer with sigmoid non-linearity is added to produce the saliency map. The weights for the decoder are randomly initialized. The final output of the network is a saliency map in the same size to input image.

The implementation details of SalGAN are presented in Table \ref{tab:generator}.

\begin{table}
\centering
\footnotesize
\begin{tabular}{lrrrrr}
\hline
layer & depth & kernel & stride & pad & activation \\
\hline
conv1\_1 & 64& $ 1 \times 1 $ & 1 & 1 & ReLU \\
conv1\_2 & 64 & $ 3 \times 3 $ & 1 & 1 & ReLU \\
\hline
pool1 &  & $ 2 \times 2 $ & 2 & 0 & - \\
\hline
conv2\_1 & 128 & $ 3 \times 3 $ & 1 & 1 & ReLU \\
conv2\_2 & 128& $ 3 \times 3 $ & 1 & 1 & ReLU \\
\hline
pool2 &  & $ 2 \times 2 $ & 2 & 0 & - \\
\hline
conv3\_1 & 256 & $ 3 \times 3 $ & 1 & 1 & ReLU \\
conv3\_2 & 256 & $ 3 \times 3 $ & 1 & 1 & ReLU \\
conv3\_3 & 256 & $ 3 \times 3 $ & 1 & 1 & ReLU \\
\hline
pool3 &  & $ 2 \times 2 $ & 2 & 0 & - \\
\hline
conv4\_1 & 512 & $ 3 \times 3 $ & 1 & 1 & ReLU \\
conv4\_2 & 512 & $ 3 \times 3 $ & 1 & 1 & ReLU \\
conv4\_3 & 512 & $ 3 \times 3 $ & 1 & 1 & ReLU \\
\hline
pool4 &  & $ 2 \times 2 $ & 2 & 0 & - \\
\hline
conv5\_1 & 512 & $ 3 \times 3 $ & 1 & 1 & ReLU \\
conv5\_2 & 512 & $ 3 \times 3 $ & 1 & 1 & ReLU \\
conv5\_3 & 512 & $ 3 \times 3 $ & 1 & 1 & ReLU \\
\hline
conv6\_1 & 512 & $ 3 \times 3 $ & 1 & 1 & ReLU \\
conv6\_2 & 512 & $ 3 \times 3 $ & 1 & 1 & ReLU \\
conv6\_3 & 512 & $ 3 \times 3 $ & 1 & 1 & ReLU \\
\hline
upsample6 &  & $ 2 \times 2 $ & 2 & 0 & - \\
\hline
conv7\_1 & 512 & $ 3 \times 3 $ & 1 & 1 & ReLU \\
conv7\_2 & 512 & $ 3 \times 3 $ & 1 & 1 & ReLU \\
conv7\_3 & 512 & $ 3 \times 3 $ & 1 & 1 & ReLU \\
\hline
upsample7 &  & $ 2 \times 2 $ & 2 & 0 & - \\
\hline
conv8\_1 & 256 & $ 3 \times 3 $ & 1 & 1 & ReLU \\
conv8\_2 & 256 & $ 3 \times 3 $ & 1 & 1 & ReLU \\
conv8\_3 & 256 & $ 3 \times 3 $ & 1 & 1 & ReLU \\
\hline
upsample8 &  & $ 2 \times 2 $ & 2 & 0 & - \\
\hline
conv9\_1 & 128 & $ 3 \times 3 $ & 1 & 1 & ReLU \\
conv9\_2 & 128 & $ 3 \times 3 $ & 1 & 1 & ReLU \\
\hline
upsample9 &  & $ 2 \times 2 $ & 2 & 0 & - \\
\hline
conv10\_1 & 64 & $ 3 \times 3 $ & 1 & 1 & ReLU \\
conv10\_2 & 64 & $ 3 \times 3 $ & 1 & 1 & ReLU \\
\hline
output&1&1x1&1&0&Sigmoid\\
\hline
\end{tabular}
\caption{Architecture of the generator network.}
\label{tab:generator}
\end{table}

\subsection{Discriminator}
\label{ssec:Discriminator}

\par Table~\ref{tab:discriminator} gives the architecture and layer configuration for the discriminator. In short, the network is composed of six 3x3 kernel convolutions interspersed with three pooling layers ($\downarrow$2), and followed by three fully connected layers. The convolution layers all use ReLU activations while the fully connected layers employ $\tanh$ activations, with the exception of the final layer, which uses a sigmoid activation. 

\begin{table}
\begin{tabular}{lrrrrr}
\hline
layer & depth & kernel & stride & pad & activation \\
\hline
conv1\_1 & 3 & $ 1 \times 1 $ & 1 & 1 & ReLU \\
conv1\_2 & 32 & $ 3 \times 3 $ & 1 & 1 & ReLU \\
pool1 &  & $ 2 \times 2 $ & 2 & 0 & - \\
\hline
conv2\_1 & 64 & $ 3 \times 3 $ & 1 & 1 & ReLU \\
conv2\_2 & 64 & $ 3 \times 3 $ & 1 & 1 & ReLU \\
pool2 &  & $ 2 \times 2 $ & 2 & 0 & - \\
\hline
conv3\_1 & 64 & $ 3 \times 3 $ & 1 & 1 & ReLU \\
conv3\_2 & 64 & $ 3 \times 3 $ & 1 & 1 & ReLU \\
pool3 &  & $ 2 \times 2 $ & 2 & 0 & - \\
\hline
fc4 & 100 & - & - & - & tanh \\
fc5 & 2 & - & - & - & tanh \\
fc6 & 1 & - & - & - & sigmoid \\
\hline
\end{tabular}
\caption{Architecture of the discriminator network.}
\label{tab:discriminator}
\end{table}

\section{Training}
\label{sec:Training}

The filter weights in SalGAN have been trained over a perceptual loss \citep{Johnson2016eccv} resulting from combining a content and adversarial loss. 
The content loss follows a classic approach in which the predicted saliency map is pixel-wise compared with the corresponding one from ground truth.
The adversarial loss depends of the real/synthetic prediction of the discriminator over the generated saliency map.

\subsection{Content loss}
\label{ssec:Content}

The content loss is computed in a per-pixel basis, where each value of the predicted saliency map is compared with its corresponding peer from the ground truth map.
Given an image $I$ of dimensions $N = W \times H$, we represent the saliency map $S$ as vector of probabilities, where $S_{j}$ is the probability of pixel $I_j$ being fixated. A content loss function $\mathcal{L}(S,\hat{S})$ is defined between the predicted saliency map $\hat{S}$ and its corresponding ground truth $S$.

The first considered content loss is mean squared error (MSE) or Euclidean loss, defined as:
\begin{equation}
\mathcal{L}_{MSE} = 
\frac{1}{N}
\sum_{j=1}^{N} (S_j - \hat{S}_j)^{2}.
\end{equation}
In our work, MSE is used as a baseline reference, as it has been adopted directly or with some variations in other state of the art solutions for visual saliency prediction \citep{Pan_2016_CVPR,mlnet2016}.

Solutions based on MSE aim at maximizing the peak signal-to-noise ratio (PSNR).
These works tend to filter high spatial frequencies in the output, favoring this way blurred contours.
MSE corresponds to computing the Euclidean distance between the predicted saliency and the ground truth.

Ground truth saliency maps are normalized so that each value is in the range $[0, 1]$. Saliency values can therefore be interpreted as estimates of the probability that a particular pixel is attended by an observer. It is tempting to therefore induce a multinomial distribution on the predictions using a softmax on the final layer. Clearly, however, more than a single pixel may be attended, making it more appropriate to treat each predicted value as independent of the others. We therefore propose to apply an element-wise sigmoid to each output in the final layer so that the pixel-wise predictions  can be thought of as probabilities for independent binary random variables. An appropriate loss in such a setting is the
binary cross entropy, which is the average of the individual binary cross entropies (BCE) across all pixels:
\begin{equation}
\mathcal{L}_{BCE} = 
-\frac{1}{N}
\sum_{j=1}^{N} 
(S_j \log(\hat{S}_j) + (1 - S_j)\log(1 - \hat{S}_j)).  
\end{equation}

\subsection{Adversarial loss}
\label{ssec:Adversarial}

Generative adversarial networks (GANs)~\citep{goodfellow2014generative} are commonly used to generate images with realistic statistical properties. The idea is to simultaneously fit two parametric functions. The first of these functions, known as the generator, is trained to transform samples from a simple distribution (e.g. Gaussian) into samples from a more complicated distribution (e.g. natural images). The second function, the discriminator, is trained to distinguish between samples from the true distribution and generated samples. Training proceeds alternating between training the discriminator using generated and real samples, and training the generator, by keeping the discriminator weights constant and backpropagating the error through the discriminator to update the generator weights.

The saliency prediction problem has some important differences from the above scenario. First, the objective is to fit a deterministic function that predict realistic saliency values from images, rather than realistic images from random noise. As such, in our case the input to the generator (saliency prediction network) is not random noise but an image. Second, the input image that a saliency map corresponds to is essential, due the fact the goal is not only to have the two saliency maps becoming indistinguishable but with the condition that they both correspond the same input image. We therefore include both the image and saliency map as inputs to the discriminator network. Finally, when using generative adversarial networks to generate realistic images, there is generally no ground truth to compare against. In our case, however, the corresponding ground truth saliency map is available. When updating the parameters of the generator function, we found that using a loss function that is a combination of the error from the discriminator and the cross entropy with respect to the ground truth improved the stability and convergence rate of the adversarial training.
The final loss function for the saliency prediction network during adversarial training can be formulated as:
\begin{equation}
\mathcal{L} =
\alpha \cdotp \mathcal{L}_{BCE} +
L( D(I,\hat{S}), 1).
\label{eq:alpha}
\end{equation}
where $L$ is the binary cross entropy loss, and 1 is the target category of real samples and 0 for the category of fake (predicted) sample.  Here,  instead of minimizing $-L( D(I,S), 0)$, we optimize  $L( D(I,S), 1)$ which provides stronger gradient, similar to \citep{goodfellow2014generative} .
$D(I,\hat{S})$ is the probability of fooling the discriminator network, so that
the loss associated to the saliency prediction network will grow more when chances of fooling the discriminator are lower.
During the training of the discriminator, no content loss is available and  the loss function is:
\begin{equation}
\mathcal{L_D} =
L( D(I,S), 1) +
L( D(I,\hat{S}), 0).
\end{equation}
At train time, we first bootstrap the saliency prediction network  function by training for 15 epochs using only BCE, which is computed with respect to the down-sampled output and ground truth saliency. After this, we add the discriminator and begin adversarial training. The input to the discriminator network is an RGBS image of size $256\times 192\times 4$ containing both the source image channels and (predicted or ground truth) saliency. 


We train the networks on the 15,000 images from the SALICON training set using a batch size of 32. 
During the adversarial training, we alternate the training of the saliency prediction network and discriminator network after each iteration (batch). We used L2 weight regularization (i.e. weight decay) when training both the generator and discriminator ($\lambda = 1\times 10^{-4}$). We used AdaGrad for optimization, with an initial learning rate of $3\times 10^{-4}$.








\section{Experiments}
\label{sec:Experiments}

\par The presented SalGAN model for visual saliency prediction was assessed and compared from different perspectives. First, the impact of using BCE and the downsampled saliency maps are assessed. Second, the gain of the adversarial loss is measured and discussed, both from a quantitative and a qualitative point of view. Finally, the performance of SalGAN is compared to published works to compare its performance with the current state-of-the-art.
The experiments aimed at finding the best configuration for SalGAN were run using the \textit{train} and \textit{validation} partitions of the SALICON dataset~\citep{jiang2015salicon}.
This is a large dataset built by collecting mouse clicks on a total of 20,000 images from the Microsoft Common Objects in Context (MS-CoCo) dataset~\citep{lin2014microsoft}.
We have adopted this dataset for our experiments because it is the largest one available for visual saliency prediction.
In addition to SALICON,we also present results on MIT300, the benchmark with the largest amount of submissions.

\begin{table}
\begin{center}
\begin{tabular}{lccccc}
\hline
 			& sAUC $\uparrow$ 			& AUC-B $\uparrow$		& NSS $\uparrow$ 	& CC $\uparrow$ 		& IG $\uparrow$\\
\hline 
BCE 		& 0.752					& 0.825						& 2.473				& 0.761  				& 0.712 \\
BCE/2 		& 0.750					& 0.820						& \textbf{2.527}	& \textbf{0.764} 		& 0.592 \\
BCE/4 		& \textbf{0.755}		& \textbf{0.831}			& 2.511				& 0.763 				& 0.825 \\
BCE/8 		& 0.754					& 0.827						& 2.503				& 0.762 				& \textbf{0.831}\\
\hline
\end{tabular}
\end{center}
\caption{Impact of downsampled saliency maps at (15 epochs) evaluated over SALICON validation. BCE/$x$ refers to a downsample factor of $1/x$ over a saliency map of $256 \times 192$.}
\label{tab:Downsample15}
\end{table}


\subsection{Non-adversarial training}
\label{ssec:ContentLossExp}

The two content losses presented in content loss section, MSE and BCE, were compared to define a baseline upon which we later assess the impact of the adversarial training.
The two first rows of Table~\ref{tab:Adversarial} shows how a simple change from MSE to BCE brings a consistent improvement in all metrics. This improvement suggests that treating saliency prediction as multiple binary classification problem is more appropriate than treating it as a standard regression problem, in spite of the fact that the target values are not binary. Minimizing cross entropy is equivalent to minimizing the KL divergence between the predicted and target distributions, which is a reasonable objective if both predictions an targets are interpreted as probabilities.


Based on the superior BCE-based loss compared with MSE, we also explored the impact of computing the content loss over downsampled versions of the saliency map.
This technique reduces the required computational resources at both training and test times and, as shown in Table~\ref{tab:Downsample15}, not only does it not decrease performance, but it can actually improve it.
Given this results, we chose to train SalGAN on saliency maps downsampled by a factor $1/4$, which in our architecture corresponds to saliency maps of $64 \times 48$.

\subsection{Adversarial gain}

\begin{figure}[ht]
\centering
\includegraphics[width=1.0\columnwidth]{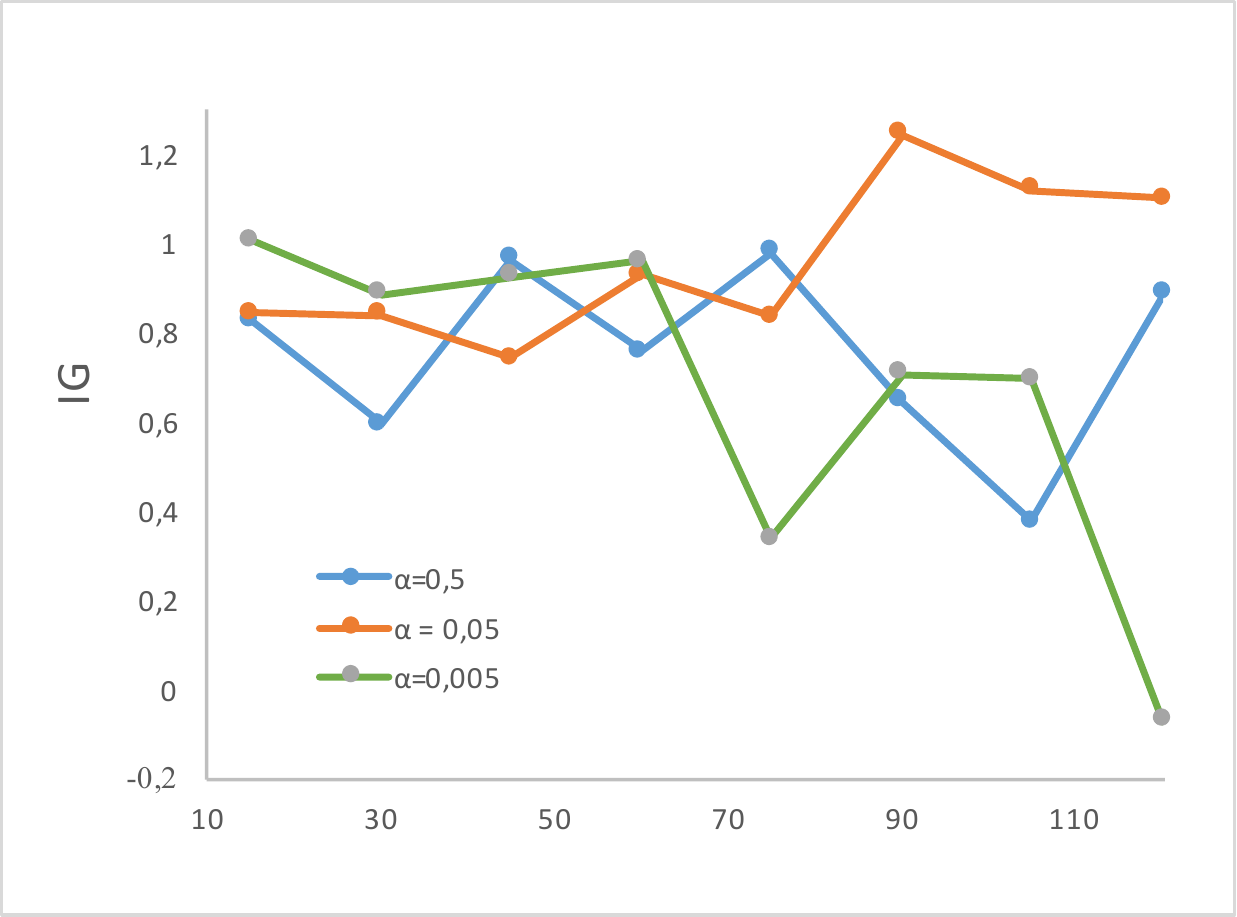}
\caption{SALICON validation set Information Gain for different hyper parameter $\alpha$ on varying numbers of epochs.}
\label{fig:alpha}
\end{figure}

The adversarial loss was introduced after estimating the value of the hyperparameter $\alpha$ in Equation \ref{eq:alpha} by maximizing the most general metric, Information Gain (IG).  As shown in Figure \ref{fig:alpha}, the search was performed on logarithmic scale, and we achieved the best performance for $\alpha=0.005$. 

The Information Gains (IG) of SalGAN  for different values of the hyper parameter $\alpha$ are compared in Figure~\ref{fig:alpha}. The search for finding an optimal hyper parameter $\alpha$ is performed on logarithmic scale, and we achieved the best performance for $\alpha=0.005$.

\begin{figure}
\centering
\includegraphics[width=1.0\columnwidth]{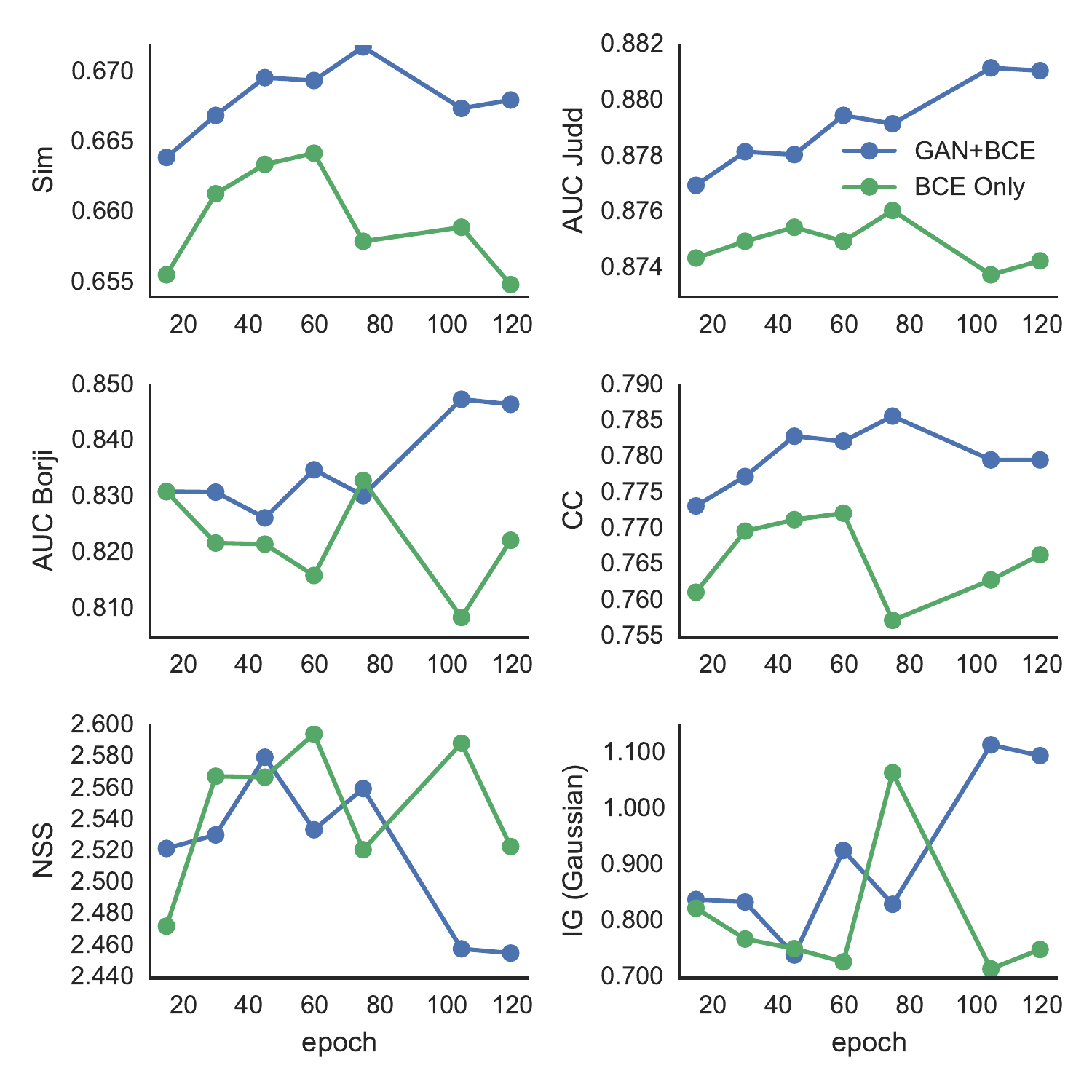}
\caption{SALICON validation set accuracy metrics for Adversarial+BCE vs BCE on varying numbers of epochs. AUC shuffled is omitted as the trend is identical to that of AUC Judd.\vspace*{-4mm}}
\label{fig:ganvsbce}
\end{figure}

The gain achieved by introducing the adversarial loss into the perceptual loss was assessed by using BCE as a content loss and feature maps of $68 \times 48$.
The first row of results in Table~\ref{tab:Adversarial} refers to a baseline defined by training SalGAN with the BCE content loss for 15 epochs only. 
Later, two options are considered: 1) training based on BCE only (2nd row), or 2) introducing the adversarial loss (3rd and 4th row). 

Figure~\ref{fig:ganvsbce} compares validation set accuracy metrics for training with combined GAN and BCE loss versus a BCE alone as the number of epochs increases. In the case of the AUC metrics (Judd and Borji), increasing the number of epochs does not lead to significant improvements when using BCE alone. The combined BCE/GAN loss however, continues to improve performance with further training. After 100 and 120 epochs, the combined GAN/BCE loss shows substantial improvements over BCE for five of six metrics.

The single metric for which Adversarial training fails to improve performance is normalized scanpath saliency (NSS). The reason for this may be that GAN training tends to produce a smoother and more spread out estimate of saliency, which better matches the statistical properties of real saliency maps, but may increase the false positive rate. As noted in \citep{Bylinskii2016metrics}, NSS is very sensitive to such false positives. The impact of increased false positives depends on the final application. In applications where the saliency map is used as a multiplicative attention model (e.g. in retrieval applications, where spatial features are importance weighted), false positives are often less important than false negatives, since while the former includes more distractors, the latter removes potentially useful features. Note also that NSS is differentiable, so could potentially be optimized directly when important for a particular application.


\begin{table}
\begin{center}
\begin{tabular}{lcccccc}
\hline
			& sAUC $\uparrow$ 	& AUC-B $\uparrow$		& NSS $\uparrow$ 	& CC $\uparrow$ 	& IG \\
\hline
MSE			& 0.728			& 0.820				& 1.680			& 0.708 			& 0.628 \\
BCE			& 0.753			& 0.825				& 2.562			& 0.772 			& 0.824 \\
\hline
BCE/4		& 0.757			& 0.833				& \textbf{2.580}	& 0.772 			& 1.067\\
GAN/4		& \textbf{0.773}	& \textbf{0.859}		& 2.560			& \textbf{0.786} 	& \textbf{1.243}\\

\hline
\end{tabular}
\end{center}
\caption{Best results through epochs obtained with non-adversarial (MSE and BCE) and adversarial training. BCE/4 and GAN/4 refer to downsampled saliency maps.  Saliency maps assessed on SALICON validation. }
\label{tab:Adversarial}
\end{table}



\subsection{Comparison with the state-of-the-art}
\label{ssec:SoAExp}


SalGAN is compared in Table~\ref{tab:soa} to several other algorithms from the state-of-the-art.
The comparison is based on the evaluations run by the organizers of the SALICON and MIT300 benchmarks on a test dataset whose ground truth is not public.
The two benchmarks offer complementary features: while SALICON is a much larger dataset with 5,000 test images, MIT300 has attracted the participation of many more researchers.
In both cases, SalGAN was trained using 15,000 images contained in the training (10,000) and validation (5,000) partitions of the SALICON dataset.
Notice that while both datasets aim at capturing visual saliency, the acquisition of data differed, as SALICON ground truth was generated based on crowdsourced mouse clicks, while the MIT300 was built with eye trackers on a limited and controlled group of users.
Table \ref{tab:soa} compares SalGAN with other contemporary works that have used SALICON and MIT300 datasets.
SalGAN presents very competitive results in both datasets, as it improves or equals the performance of all other models in at least one metric.

\begin{table*}
\small
\begin{center}
\begin{tabular}{lcccccccc}
\hline
SALICON (test)				&	AUC-J $\uparrow$	& Sim $\uparrow$ & EMD  $\downarrow$ & AUC-B $\uparrow$ & sAUC $\uparrow$ 	& CC $\uparrow$ & NSS $\uparrow$ & KL $\downarrow$ \\		
\hline
DSCLRCN \citep{liu2018deep} 			& -					&	-			&	-				&	0.884			&	0.776			&	0.831		& 3.157			&	-				\\

\textbf{SalGAN}				& -						&	-			&	-				&	\textbf{0.884}	&	\textbf{0.772}	&	\textbf{0.781}	& \textbf{2.459} &	-				\\
ML-NET \citep{mlnet2016}		& -						&	-			&	-				&	(0.866)			&	(0.768)			&	(0.743)		& 2.789			&	-				\\
SalNet \citep{Pan_2016_CVPR}	& -						&	-			&	-				&	(0.858)			&	(0.724)			&	(0.609)		& (1.859)		&	-				\\
\hline
MIT300						&	AUC-J $\uparrow$	& Sim $\uparrow$ & EMD  $\downarrow$ & AUC-B $\uparrow$ & sAUC $\uparrow$ 	& CC $\uparrow$ & NSS $\uparrow$ & KL $\downarrow$ \\
\hline
Humans 						& 0.92	 	& 1.00		& 0.00 		& 0.88		& 0.81 		& 1.0 		& 3.29 		& 0.00	\\
Deep Gaze II \cite{kummerer2017understanding}			& (0.84)		& (0.43)	& (4.52)	& (0.83)		& 0.77		& (0.45)	& (1.16) 	& (1.04) \\
DSCLRCN \citep{liu2018deep}	& 0.87		& 0.68		& 2.17		& (0.79)	& 0.72		& 0.80		& 2.35		& 0.95\\	
SALICON	\citep{huang2015salicon}		& 0.87		& (0.60)	& (2.62)	& 0.85		& 0.74		& 0.74		& 2.12		& 0.54 \\
\textbf{SalGAN}				& \textbf{0.86}	& \textbf{0.63}		& \textbf{2.29}		& \textbf{0.81}		& \textbf{0.72}		& \textbf{0.73}	& \textbf{2.04}	& \textbf{1.07} \\
PDP	\citep{jetley2016end}	& (0.85)	& (0.60)	& (2.58)	& (0.80)	& 0.73		& (0.70)	& 2.05		& 0.92 \\
ML-NET \citep{mlnet2016}		& (0.85)	& (0.59)	& (2.63)	& (0.75)	& (0.70)	& (0.67)	& 2.05		& (1.10) \\
Deep Gaze I \citep{kummerer2015deep} & (0.84)& (0.39) & (4.97)	& 0.83		& (0.66)	& (0.48)	& (1.22)	& (1.23) \\
SalNet	\citep{Pan_2016_CVPR}& (0.83)		& (0.52) & (3.31)	& 0.82		& (0.69)	& (0.58)	& (1.51)	& 0.81 \\
BMS \citep{Zhang2013iccv}	& (0.83)	& (0.51)	& (3.35)	& 0.82		& (0.65)	& (0.55)	& (1.41)	& 0.81 \\ 			
\hline
\end{tabular}
\end{center}
\caption{Comparison of SalGAN with other state-of-the-art solutions on the SALICON (test) and MIT300 benchmarks. Values in brackets correspond to performances worse than SalGAN.}
\label{tab:soa}
\end{table*}

\subsection{Qualitative results}

The impact of adversarial training has also been explored from a qualitative perspective by observing the resulting saliency maps.
\begin{figure}[t]
\includegraphics[width=\linewidth]{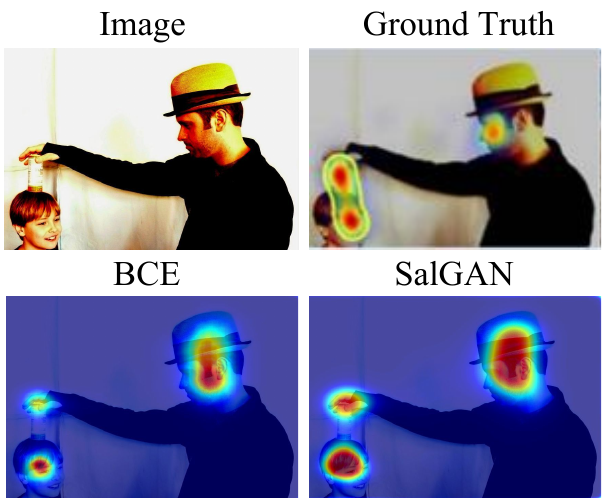}
\caption{Example images from MIT300 containing salient region (marked in yellow) that is often missed by computational models, and saliency map estimated by SalGAN.}
\label{fig:MITExamples}
\end{figure}
Figure~\ref{fig:MITExamples} shows one example from the MIT300 dataset, highlighted in \citep{Bylinsk2016eccv} as being particular challenges for existing saliency algorithms. The areas highlighted in yellow in the images on the left are regions that are typically missed by algorithms. In the this  example, we see that SalGAN successfully detects the often missed hand of the magician and face of the boy as being salient. 

\begin{figure}
\centering
\includegraphics[width=1.0\columnwidth]{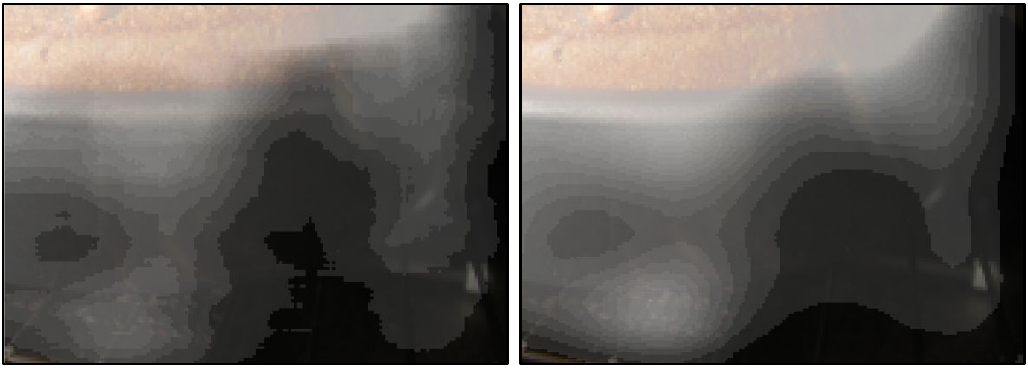}
\caption{Close-up comparison of output from training on BCE loss vs combined BCE/Adversarial loss. Left: saliency map from network trained with BCE loss. Right: saliency map from proposed adversarial training.}
\label{fig:bcsgan_artifacts}
\end{figure}

\par Figure~\ref{fig:bcsgan_artifacts} illustrates the effect of adversarial training on the statistical properties of the generated saliency maps. Shown are two close up sections of a saliency map from cross entropy training (left) and adversarial training (right). Training on BCE alone produces saliency maps that while they may be locally consistent with the ground truth, are often less smooth and have complex level sets. Adversarial training on the other hand produces much smoother and simpler level sets.
\begin{figure*}
\centering
\includegraphics[width=1\linewidth]{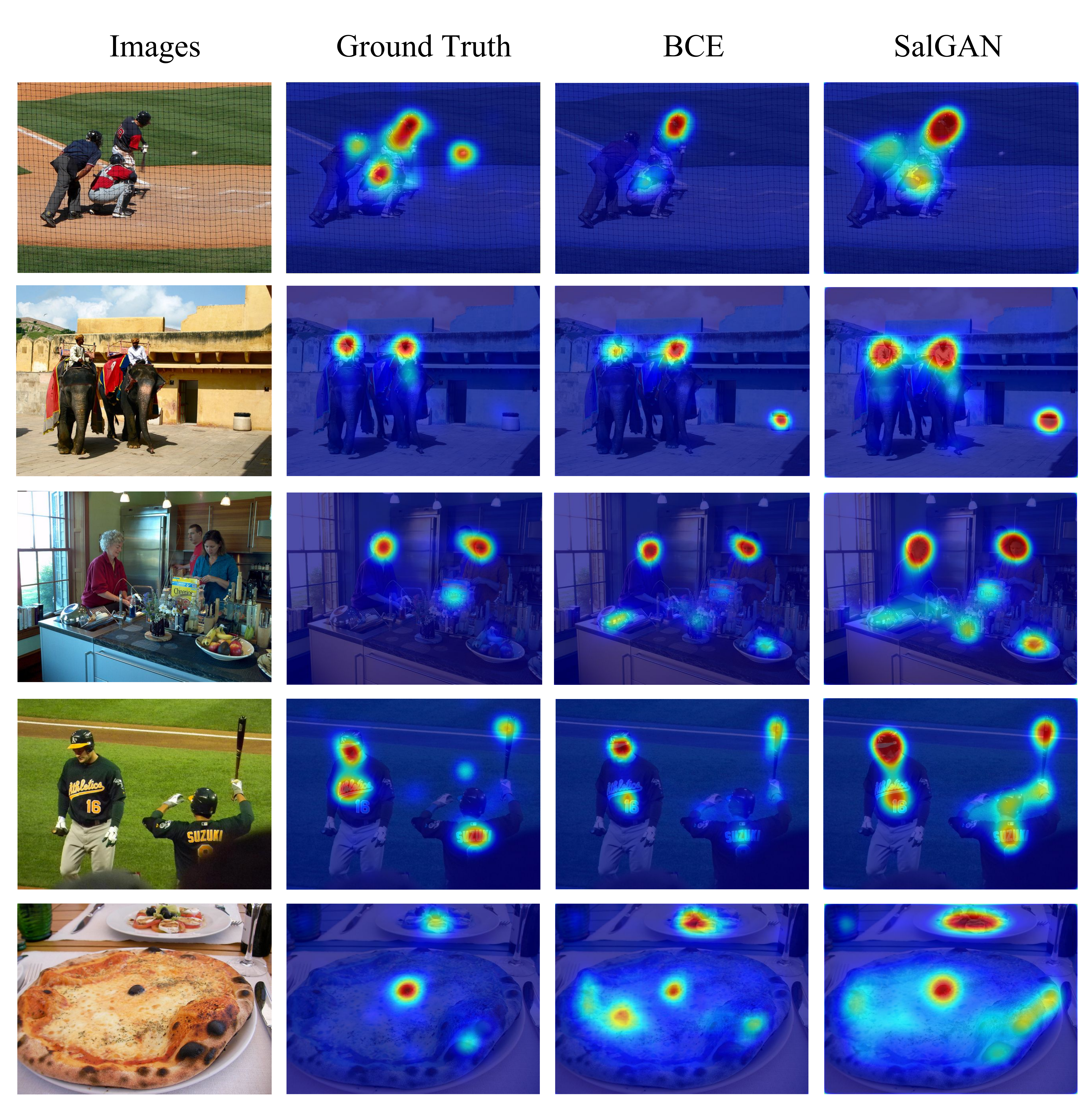}
\caption{Qualitative results of SalGAN on the SALICON validation set. SalGAN predicts well those high salient regions which are missed by BCE model. Saliency maps of BCE model are very localized in a few salient regions, they tend to fail when the number of salient regions increases. }
\label{fig:qualitativeResults}
\end{figure*}
Finally, Figure~\ref{fig:qualitativeResults} shows some qualitative results comparing the results from training with BCE and BCE/Adversarial against the ground truth for images from the SALICON validation set. 

\section{Conclusions}
\label{sec:Conclusions}

To the best of our knowledge, this is the first work that proposes an adversarial-based approach to saliency prediction and has shown how adversarial training over a deep convolutional neural network can achieve state-of-the-art performance with a simple encoder-decoder architecture.
A BCE-based content loss was shown to be effective for both initializing the saliency prediction network, and as a regularization term for stabilizing adversarial training. Our experiments showed that adversarial training improved all bar one saliency metric when compared to further training on cross entropy alone.

It is worth pointing out that although we use a VGG-16 based encoder-decoder model as the saliency prediction network in this paper, the proposed adversarial training approach is generic and could be applied to improve the performance of other saliency models. 



\section*{Acknowledgments}
The Image Processing Group at UPC is supported by the project TEC2016-75976-R, funded by the Spanish Ministerio de Economia y Competitividad and the European Regional Development Fund (ERDF). 
This material is based upon works supported by Science Foundation Ireland under Grant No 15/SIRG/3283.
We gratefully acknowledge the support of NVIDIA Corporation for the donation of GPUs used in this work.





\bibliographystyle{model2-names}
\bibliography{refs}



\end{document}